\documentclass[sigconf]{acmart}
\renewcommand\footnotetextcopyrightpermission[1]{} 
\usepackage{fancyhdr}
\AtBeginDocument{%
  \providecommand\BibTeX{{%
    \normalfont B\kern-0.5em{\scshape i\kern-0.25em b}\kern-0.8em\TeX}}}

\usepackage{enumerate}
\usepackage{anyfontsize}

\usepackage{algorithm}
\usepackage{algpseudocode}
\usepackage{amsmath}
\usepackage{subfigure}
\usepackage{epstopdf}

\begin{document}

\title{Auto-CASH: Autonomous Classification Algorithm Selection with Deep Q-Network}

\author{Tianyu Mu}
\authornotemark[1]
\email{mutianyu@hit.edu.cn}
\affiliation{%
  \institution{Harbin Institute of Technology}
  \department{Department of Computer Science}
  \state{Harbin}
  \country{China}
}

\author{Hongzhi Wang}
\authornotemark[2]
\email{wangzh@hit.edu.cn}
\affiliation{%
  \institution{Harbin Institute of Technology}
  \department{Department of Computer Science}
  \state{Harbin}
  \country{China}
}

\author{Chunnan Wang}
\authornotemark[3]
\email{WangChunnan@hit.edu.cn}
\affiliation{%
  \institution{Harbin Institute of Technology}
  \department{Department of Computer Science}
  \state{Harbin}
  \country{China}
}

\author{Zheng Liang}
\authornotemark[4]
\email{lz20@hit.edu.cn}
\affiliation{%
  \institution{Harbin Institute of Technology}
  \department{Department of Computer Science}
  \state{Harbin}
  \country{China}
}








\begin{abstract}
The great amount of datasets generated by various data sources have posed the challenge to machine learning algorithm selection and hyperparameter configuration. For a specific machine learning task, it usually takes domain experts plenty of time to select an appropriate algorithm and configure its hyperparameters. If the problem of algorithm selection and hyperparameter optimization can be solved automatically, the task will be executed more efficiently with performance guarantee. Such problem is also known as the CASH problem. Early work either requires a large amount of human labor, or suffers from high time or space complexity. In our work, we present Auto-CASH, a pre-trained model based on meta-learning, to solve the CASH problem more efficiently. Auto-CASH is the first approach that utilizes Deep Q-Network to automatically select the meta-features for each dataset, thus reducing the time cost tremendously without introducing too much human labor. To demonstrate the effectiveness of our model, we conduct extensive experiments on 120 real-world classification datasets. Compared with classical and the state-of-art CASH approaches, experimental results show that Auto-CASH achieves better performance within shorter time.
\end{abstract}



\keywords{Meta-feature, Algorithm selection, Hyperparameter optimization, CASH problem, Classification algorithm, Deep Q-Network}


\maketitle

\section{Introduction}
Machine learning(ML) approaches have been used widely in recent years to solve problems in the data science field~\cite{xiao2017fashion}, such as data mining, data preprocessing, etc. Many algorithms(or models)have been developed for a specific problem~\cite{lindauer2019algorithm,taylor2018adaptive}. However, for different datasets, the performance of these algorithms varies considerably. One learning algorithm cannot outperform another learning algorithm in various aspects and problems~\cite{schaffer1994cross}. Therefore, domain experts usually choose the most suitable algorithm and hyperparameters based on their experience and a series of experiments to optimize performance the.

However, with the explosive growth of both data and ML algorithms, the former approach could hardly work. Each algorithm has a large hyperparameter configuration space. Even for an expert with adequate domain knowledge, it will be hard to make an ideal selection among various algorithms and their complex hyperparameter space. In the face of such situation, Thornton et al. presented the combined algorithm selection and hyperparameter optimization problem(CASH)~\cite{thornton2013auto}, aiming at helping other researchers find a solution to select a suitable algorithm and configure the hyperparameters in different scenarios automatically.

An effective approach to solve the CASH problem is {\itshape meta-learning}, also known as {\itshape learn how to learn}. With meta-feature vector representing the previous experience, the meta-learning is capable of recommending the same algorithm for similar tasks~\cite{hutter2019automated,lake2017building}. Meta-learning requires less human labor and computation resources, making it more suitable for the automatic and lightweight demand in practice.

Therefore, to solve the CASH problem in an automatic and lightweight way, there are two main challenges. On the one hand, we should make the whole workflow automatically. An effective strategy should be determined to automatically choose the meta-feature used. The correlations among meta-feature candidates are complicated, and their influence on the algorithm selection result is inexplicable, which makes it crucial to select the optimal meta feature. On the other hand, CASH has buckets effect. That is, the measurement of HPO results has multi-aspect on real-world task, and the usability depends on the shortest aspect. The HPO algorithm adopted should have performance guarantee, acceptable time cost and the potential to deal with various data types.

Auto-WEKA~\cite{thornton2013auto} is the first approach which provides a solution to the CASH problem. It uses a hyperparameter to represent candidate algorithms, thereby converting the CASH problem into a hyperparameter optimization problem(HPO). However, Auto-WEKA will iterate online round by round to find the best solution, thus suffering from high time and space cost. Different from Auto-WEKA, Auto-Model\cite{wang2019auto} extracts experimental results from previously published ML papers to create a knowledge base, making the selection of algorithms more intelligent and automated. The knowledge base can be updated with continuous training. A steady flow of training data will enhance the knowledge base gradually replacing the experience of experts. To the best of our knowledge, Auto-Model performs better than Auto-WEKA on classification problems. Nevertheless, the quality of used paper will affect the effectiveness of the entire model, too much manual work is need for evaluating each paper's contribution to the knowledge base. As a consequence, Auto-Model is not a fully automated CASH processing model.

From above discussions, early works\footnote{Auto-Model and Auto-WEKA} cannot solve those challenges well, which makes them inefficient in practice.
Thus, we present {\itshape Auto-CASH}, a pre-trained model based on meta-learning, to slove the CASH problem in an efficient way. For the first challenge, Auto-CASH utilizes {\itshape Deep Q-Network}\cite{mnih2013playing}, a reinforcement learning(RL) approach, to automatically select meta-feature. Then given each training dataset, we use its meta-feature~\cite{bilalli2017predictive,filchenkov2015datasets}, along with the most suitable algorithm tested for it, to train a {\itshape Random Forest} (RF) classifier, which is the key to the algorithm selection process~\footnote{The prediction function of a trained RF can infer the most suitable algorithm for a new task instance.}. By RF, Auto-CASH achieve a good performance and an acceptable time cost. For the second challenge, we adopt {\itshape Genetic Algorithm} (GA), which is one of the fastest and the most effective HPO approaches to improve the efficiency of finding the optimal hyperparameter setting. Our experimental results show that GA spends a quarter less time on HPO than early work and achieves better results.

Major contributions in our work are summarized as follows:
\begin{enumerate}[1)]
  \item We propose {\itshape Auto-CASH}, a meta-learning based approach for the CASH problem. By sufficiently utilizing the experiences of training datasets, our approach is more lightweight and efficient in practice.
  \item We first transform the selection of meta-feature into a continuous action decision problem. Deep Q-Network is introduced to automatically choose the meta-features we use in the algorithm selection process. To the best of our knowledge, Auto-CASH is the first study that introduce RL approach and meta-learning to the CASH problem.
  \item We conduct extensive experiments and demonstrate the effectiveness of Auto-CASH on 120 real-world classification datasets from UCI Machine Learning Repository~\footnote{https://archive.ics.uci.edu/ml/index.php} and Kaggle~\footnote{https://www.kaggle.com/}. Compared with Auto-Model and Auto-WEKA, experimental results show that Auto-CASH can deal with the CASH problem better.
\end{enumerate}

The structure of this paper is as follows. Section \ref{sec: pre} introduces some concepts of DQN and GA, which are crucial in Auto-CASH. Section \ref{sec:overview} describes the workflow and some implementations of our model. Section \ref{sec: methodology} introduces the methodology for automatically meta-feature selection. Section \ref{sec: eval} evaluates the performance of Auto-CASH and compares it with early work. Finally, we draw a conclusion and give our future research directions in section \ref{sec: conclusion}.

\section{PREREQUISITES}
\label{sec: pre}
In this section, we introduce the basic concepts of Deep Q-Network and HPO algorithm, respectively. Both of them are crucial in Auto-CASH.

\subsection{Deep Q-Network}
In the first part of Auto-CASH, the selection of meta-feature is transformed into a continuous action decision problem, which can be solved by RL approaches. A RL approach includes two entities: the agent and the environment. The interactions between the two entities are as follows. Under the state $s_{t}$, the agent takes an action $a_{t}$ and get a corresponding reward $r_{t}$ from the environment, and then the agent enters next state $s_{t+1}$. The action decision process will be repeated until the agent meets termination conditions.

Continuous action decision problems have the following characteristics.
\begin{itemize}
  \item For various actions, the corresponding rewards are usually different.
  \item Reward for an action is delayed.
  \item Value of the reward for an action is influenced by the current state.
\end{itemize}

Q-learning\cite{watkins1992q} is a classical value-based RL algorithm to solve continuous action decision problems. Let $Q(s_{t}, a_{t})$ value represents the reward of action $a_{t}$ in state $s_{t}$. The main idea of Q-learning is to fill in the Q-table with $Q(s_{t} \in all States, a_{t} \in all Actions)$ value by iterative training. At the beginning of the training phase, i.e. exploring the environment, the Q-table is filled up with the same random initial value. As the agent explores the environment continuously, the Q-table will be updated using Equation (\ref{equation 1}).
\begin{equation}
\label{equation 1}
Q(s,a) \leftarrow Q(s, a) + \alpha[r + \gamma max_{a'}Q(s',a')-Q(s,a)]
\end{equation}

\begin{figure}[h]
  \centering
  \includegraphics[width=\linewidth]{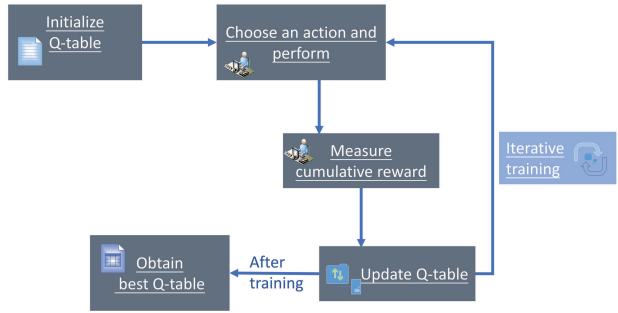}
  \caption{Q-learning workflow.}
  \Description{111111111111111111111111}
  \label{q-learning pic}
\end{figure}

Under the state $s$, the agent select action $a$ which can obtain a maximum cumulative reward $r$ according to Q-table, then  enters state $s'$. The Q-table should be updated right now. $\gamma, \alpha$ denote the discount factor and learning rate, respectively. The difference between the true Q-value and the estimated Q-value is $\alpha[R(s,a) + \gamma max_{a'}Q(s',a')-Q(s,a)]$\cite{melo2001convergence}. The value of $\alpha$ determines the speed of learning new informations and the value of $\gamma$ means the importance of future rewards. The specific execution workflow of Q-learning is shown in Figure \ref{q-learning pic}.

Q-learning utilizes tables to store each state and the Q value of each action under this state. However, with the problem getting complicated, it is difficult to describe the environment by an acceptable amount of states an agent could possibly enter. If we still use Q-table, there should be heavy space cost. Searching in such a complex table also needs a lot of time and computing resources.   Deep Q-Network (DQN)\cite{mnih2013playing} is proposed, which uses neural network (NN) to analyze the reward of each action under a specific state instead of Q-table.  The input of DQN is state values and the output is the estimated reward value for each action. The agent then randomly chooses actions with a probability of $\varepsilon  (0 < \varepsilon < 1)$ and chooses actions with a probability of $1 - \varepsilon$ that can bring the maximal reward. It is called $\varepsilon - greedy$ exploration strategy, which can balance the exploration and the exploitation. In the beginning, the system will maximize the exploration space completely randomly. As training continues, $\varepsilon$ will gradually decrease from 1 to 0. Finally it will be fixed to a stable exploration rate.

In the training phase, DQN uses the same strategy with Q-learning to update the parameter values of NN. Besides, DQN has two mechanisms to make it acts like a human being: {\itshape Experience Replay} and {\itshape Fixed Q-target}.  Every time the DQN is updated, we can randomly extract some previous experiences stored in the experience base to learn. Randomly extracting disrupts the correlation between experiences and makes update process more efficient. Fixed Q-targets is also a mechanism that disrupts correlations. We use two NNs with the same structure but different parameters to predict the estimated and target Q-value, respectively.  NN for estimated Q-value has the latest parameter values, while for target Q-value, it has previous parameters. With these two mechanisms, DQN becomes more intelligent.

\subsection{Genetic Algorithms for HPO}

\subsubsection{Concepts of HPO}
Let $\mathcal{A}$ and $\mathcal{D}$ repersents a learning algorithm with $n$ hyperparameters $\lambda_{1}, \lambda_{2}, ... \lambda_{n}$ and a dataset, respectively. The domain of $\lambda_{i}$ is denoted by $\Lambda_{i}$. So the overall hyperparameter space $\Lambda$ is a subset of Cartesian product of these domains: $\Lambda \subseteq \Lambda_{1} \times \Lambda_{2} \times ... \Lambda_{n}$. Given a score function $\mathcal{F}(\mathcal{A}, \mathcal{D})$, the HPO problem can be written as Equation (\ref{equation 2}), where $\mathcal{A}_{\lambda}$ means algorithm $\mathcal{A}$ with a hyperparameter configuration $\lambda$ ($\lambda \in \Lambda$).
\begin{equation}
\label{equation 2}
\lambda^{*} = \mathop{\arg\max}_{\lambda \in \Lambda} \mathcal{F}(\mathcal{A}_{\lambda}, \mathcal{D})
\end{equation}

\subsubsection{Introduction to Genetic Algorithm}
Mainstream modern  ML and deep learning algorithms or models' performances are sensitive to their hyperparameter settings. To solve HPO efficiently and automatically, some classical approaches are proposed: Grid Search\cite{montgomery2017design}, Random Search\cite{bergstra2012random},  Hyperband\cite{li2017hyperband}, Bayesian Optimization\cite{pelikan1999boa,brochu2010tutorial}, and Genetic Algorithm\cite{whitley1994genetic}, etc. Among them the most famous and effective HPO approaches are Bayesian Optimization (BO)\cite{snoek2015scalable,snoek2012practical,dahl2013improving} and Genetic Algorithm (GA)\cite{zames1981genetic,olson2019tpot}.

BO is a black-box global optimization approach that almost has the best performance among the above-mentioned HPO approaches. It uses a surrogate model(eg. Gaussian Process) to fit the target function, then predict the distribution of the surrogate model based on Bayesian theory iteratively. Finally, BO returns the best result it explored as the HPO solution. However, it is time-consuming to explore the surrogate model using Bayesian theory and historical data. When encountering a model or algorithm with high time complexity or high dimensional hyperparameter space, although BO can provide the optimal HPO results, the execution time is hardly unacceptable. So in Auto-CASH, we use GA, another HPO approach with similar performance and reduce the time cost.

GA originates from the computer simulation study of biological systems. It is a stochastic global search and optimization method developed by imitating the biological evolution mechanism of nature. In essence, it is an efficient, parallel, and global search method, which automatically accumulates knowledge about the search space during the search process.

%

\section{Overview}
\label{sec:overview}

In the era of algorithms and data explosion, it is increasingly challenging to select the algorithm most suitable for different datasets in a particular task (e.g. classification). One of the best ways to solve such problems is to train a pre-model based on previous experience. In our work, we use the training dataset and its optimal algorithm as the previous experience, which is the most intuitive form of experience and easy to apply. After the training, the pre-model is like an expert who has learned all the previous experiences and can work effectively offline.

When selecting the optimal algorithm for training datasets, the metric criteria are crucial. For classification algorithms, the most commonly used metric criterion is accuracy. However, in some cases (e.g. unbalanced classification), higher accuracy does not mean better performance. A more balanced metric criterion should be considered to measure the performance of the algorithm on the dataset from multiple perspectives. Combining {\itshape accuracy} and {\itshape AUC} (area under ROC curve), we also propose a more comprehensive metric criterion based on multi-objective optimization.

In our approach, we use DQN to select meta-features representing the whole training dataset. To develop the RL environment for DQN, we need to define the reward for the action of meta-feature selection. We randomly select batch of meta-features to construct a RF, then we test its performance on the training datasets. By repeating this procedure, we can estimate the influence of each meta-feature on the classification algorithm recommended by RF as the reward of the DQN environment.

In this sections, we will first introduce the workflow of Auto-CASH in Section~\ref{sec: workflow}. Then we discuss the criterion used in Auto-CASH and its advantage in Section~\ref{sec: metric}. Eventually, we give the implementations of algorithm selection and HPO in Section~\ref{sec: algorithm selection } and Section~\ref{sec: hpo}, respectively.

\subsection{Workflow}
\label{sec: workflow}
\begin{figure}[h]
  \centering
  \includegraphics[width=\linewidth]{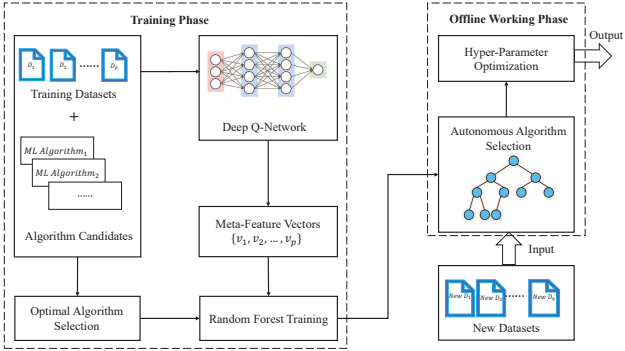}
  \caption{The Auto-CASH workflow. After training, our model can work offline.}
  \Description{4444444444}
  \label{Auto-CASH pic}
\end{figure}

  \begin{algorithm}[h]
  \caption{ Auto-CASH approach.}
  \label{alg:auto-cash}
  \begin{algorithmic}[1]
    \Require
      The training datasets $D_{train}$;
      The candidate algorithm set $Alg$;
      The datasets needs for autonomous algorithm selection and HPO $D$;
    \Ensure
      An optimal algorithm $alg$ and its hyperparameter setting $\lambda_{alg}$;
    \State Select the optimal algorithm in $Alg$ for $D_{train}$;
    \label{code: select alg}
    \State Use DQN to select meta-feature list according to $D_{train}$;
    \label{code: select optimal meta-feature}
    \State Input the meta-vector of $D_{train}$ and optimal algorithm to RF;
    \label{code: input to rf}
    \State Train the RF;
    \label{code:train rf}
    \State Utilize the trained RF to predict $alg$ for $D$;
    \label{code: algorithm selection}
    \State Utilize Genetic Algorithm to search the $\lambda_{alg}$;
    \label{code:genetic for hpo} \\
    \Return $alg, \lambda_{alg}$;
    \label{return result}
  \end{algorithmic}
\end{algorithm}

The workflow of our Auto-CASH approach is shown in Figure \ref{Auto-CASH pic} and Algorithm \ref{alg:auto-cash}. The whole workflow is divided into two phases - the training phase and offline working phase. First of all, in Line~\ref{code: select alg} of Algorithm~\ref{alg:auto-cash}, we select the optimal candidate algorithm using our new metric criterion. In the next place, we use DQN to automatically determine the meta-feature for representing datasets, as shown in Line~\ref{code: select optimal meta-feature}. In this way, refer Line~\ref{code: input to rf} and \ref{code:train rf}, the training datasets are transformed into meta-feature vectors, together with their optimal algorithm, which are used to train a RF. Given a meta-feature vector for a new dataset, the trained RF can predict the label for it (autonomous algorithm selection), which is shown in Line~\ref{code: algorithm selection}. Eventually, in Line~\ref{code:genetic for hpo}, we apply the Genetic Algorithm to search for the optimal hyperparameter configuration.

To fairly demonstrate Auto-CASH, we will come to some critical concepts. The notations of these concepts are summarized in table \ref{tab:nota and mean}.

\begin{table}
  \caption{Notations and their meanings.}
  \label{tab:nota and mean}
  \begin{tabular}{cl}
    \toprule
    Notation&Meaning\\
    \midrule
    $Alg$& Algorithm candidates list\\
    $alg$& An algorithm in $Alg$\\
    $D_{train}$ & All training datasets\\
    $MF$& Meta-feature candidates list\\
    $mf$& A meta-feature in $MF$\\
    $M_{list}$& Eventually optimal meta-feature list\\
  \bottomrule
\end{tabular}
\end{table}

\subsection{Metric Criterion}
\label{sec: metric}

AUC is the area under the ROC~\cite{powers2011evaluation} curve. For an unbalanced distributed dataset, the AUC value represents the classifier's ability to classify positive and negative examples~\cite{fawcett2006introduction}. While selecting the optimal algorithm for each training dataset, the common evaluation is to use the accuracy, which is highly influenced by the test-train splitting. To eliminate such influence, we use a score function combining AUC and accuracy.

The accuracy and AUC of a classification algorithm usually turn out to be conflicted on an unbalanced dataset. For example, in a cancer dataset, there may be only $1\%$  of cancer records(negative case). If a classifier divides all records into positive cases, the accuracy value is $0.99$, but the AUC value is only $0.5$. Therefore, optimizing both accuracy and AUC can be treated as a multi-objective optimization problem.

A classic multi-objective optimization method~\cite{xiujuan2004overview} is weighted sum, shown as $F_{score} = w_{1} \cdot accuracy + w_{2} \cdot AUC$ in our problem. However, it needs more complicated calculations to optimize the accuracy and AUC separately, and set a reasonable weight coefficient. We use a concise way to represent the score function in Auto-CASH, shown as Equation (\ref{equation 3}).
\begin{equation}
\label{equation 3}
F_{score} = accuracy \cdot AUC
\end{equation}

\subsection{Autonomous Algorithm Selection}
\label{sec: algorithm selection }

A RF model is used for autonomous algorithm selection process, which has two advantages. First, we use some complex meta-features to represent the datasets. RF is sensitive to the internal influences among these meta-features when training. Second, RF has a high prediction accuracy without the need of hyperparameter tuning.

The trained RF contains the knowledge of previous experience, which can work offline. For a new dataset, RF will recommend an algorithm, which has the best performance with high possibility. In this way, the autonomous algorithm selection process only cost a few seconds. Training a RF needs much less human labor than Auto-Model, for Auto-Mudel has to extract rules in published papers. We compare the RF with other famous classification models (e.g., KNN, SVM) in our experiments, and the results in section \ref{sec: eval} show that it is the most effective.

\subsection{HPO}
\label{sec: hpo}

 Genetic algorithm is used for HPO process. Since Auto-WEKA has a complicated hyparameter space, and HPO is the major step, the first thing considered is the number of hyperparameters for each $alg \in Alg$. We utilize GA to tune the hyperparameters for each $alg$ and determine which hyperparameters will be tuned in the HPO according to the performance improvement after tuning. According to the Occam's razor principle, in order to reduce the complexity of the algorithm of the HPO, we only select the hyperparameters that will bring a relatively large effect improvement for tuning.


\begin{figure}[h]
  \centering
  \includegraphics[width=\linewidth]{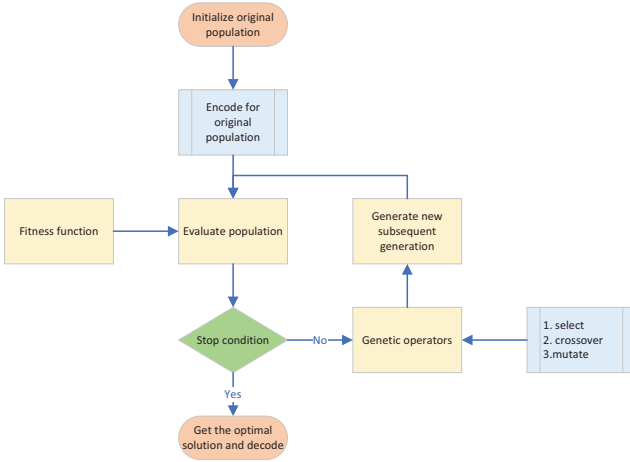}
  \caption{Genetic Algorithm workflow.}
  \Description{2222222222222222222}
  \label{GA pic}
\end{figure}
\begin{figure}[h]
  \centering
  \includegraphics[width=\linewidth]{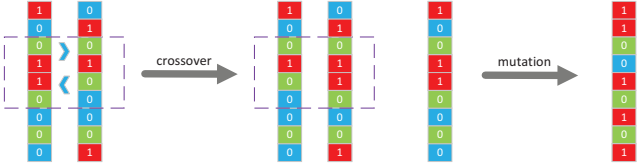}
  \caption{Crossover and mutation examples}
  \Description{33333333333}
  \label{CM pic}
\end{figure}

The workflow of GA is shown in Figure \ref{GA pic}. In the beginning, we uses binary code to encode hyperparameters and initializes the original population. Then, we select the batch of individuals with the best fitness, i.e.the algorithm performence with specific hyperparameter configuration, for subsequent generation. To introduce random disturbance, we adapt {\itshape crossover} and {\itshape mutation} as genetic operators shown in Figure \ref{CM pic}. Two binary sequences (individuals) randomly exchange their subsequences in the same position to represent the crossover process. And the binary digits of individuals alters randomly as mutation. For each subsequent generation, the hyperparameter configuration is returned as HPO result if the termination condition has been reached; otherwise, the above steps will be iteratively executed. Our experimental results in section \ref{sec: eval} show that the fitness of individuals will converge to the optimal value within 50 generations in most cases.

\section{Meta-feature selection}
\label{sec: methodology}

The major interfering factor of the algorithm selection process is the quality of meta-features. Unfortunately, due to the fact that meta-features have complicated correlation between each other, it is difficult to reconfigure the priority of them after a specific action of candidate selection. A well-studied approach focusing on the influence of multiple candidate selection is DQN. However, DQN is used to solve the automatic continuous decision problem, so we transform meta-feature selection into such problem. In the next of this section, we will discuss the methodology of DQN environment construction and using DQN to select meta-features.

First of all, we will introduce the elements of DQN, i.e. the state, action, and reward in the environment, respectively.

\textbf{Definition 1 } Given a collection of candidate meta-features $MF (|MF| = m)$, the state $s$ is the meta-features selected from $MF$. Each action $a$ selects a specific meta-feature $mf \in MF$. The eventually selected meta-features construct an optimal meta-feature list $M_{list} (M_{list} \subsetneqq MF, |M_{list}|_{max} = n, n < m)$. The reward $r_{a}$ of action $a$ is the probability of selecting the optimal algorithm by performing action $a$.

\begin{figure}[h]
  \centering
  \includegraphics[width=\linewidth]{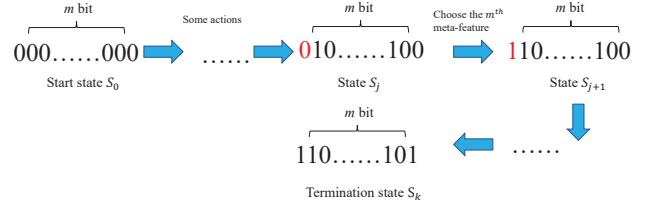}
  \caption{Transition between states examples}
  \Description{55555555555}
  \label{states pic}
\end{figure}

In Auto-CASH, we use an $m$-bit binary number to encode all states. Each bit represents a meta-feature in $MF$.  In a specific state $s$, if the meta-feature $mf$ is selected, its corresponding bit is encoded as 1. Otherwise, it is encoded as 0. Thus, there are totally $2^{m}$ states and $m$ actions. The example in Figure \ref{states pic} can explain the transition between states more clearly.
Under the start state $S_{0}$, no meta-feature has been selected, so all $m$ bits are $0$. After performing some actions, it is state $S_{j}$ now. The next action is to choose $mf_{m}$, so the $m$ th bit of the number is set $1$. These steps will be repeated until the termination state.

In order to make sufficient preparation for the RL environment, we consider several characteristics for a classification dataset to form some types of meta-fatures. For category attributes, we concentrate on the inter-class dispersion degree and the maximum range of class proportion. As for numeric attributes, we are more concerned about the center and extent of fluctuation. Besides, we also take the global numeric information of records and attributes into consider. $5$ basic types meta-features are as follows.

\begin{itemize}
  \item{\verb|Type 1|}: Category information entropy.
  \item{\verb|Type 2|}: Proportion of classes in different type of attributes.
  \item{\verb|Type 3|}: Average value.
  \item{\verb|Type 4|}: Variance.
  \item{\verb|Type 5|}: Number of instances.
\end{itemize}

On the basis of above discussion, we construct the $MF$, made up of constrained(e.g., class number in category attribute with the least classes) and combined(e.g., variance of average value in numeric attributes) meta-features. The details are shown in Section~\ref{sec: eval}.

There is no precise approach to measure or calculate the reward of each meta-feature. Therefore, we can only estimate these rewards according to experimental results on training datasets. The meta-features have complicated influence on one another, so evaluating the reward of a single meta-feature independently is not persuasive. Therefore, for each meta-feature, we randomly select some batches of meta-features containing it. With each batch of meta-features, we construct an RF. We repeat above steps multiple times for each batch size, and the average accuracy of RF is the reward. 

\begin{algorithm}[h]
  \caption{ Automatically meta-feature selection approach.}
  \label{alg:mfselection}
  \begin{algorithmic}[1]
    \Require
      The meta-feature candidates list $MF(|MF| = m)$;
      The limit of optimal meta-feature list $|M_{list}|_{max} = n(n < m)$;
      The limit of episode $e_{max}$;
    \Ensure
      The optimal meta-feature list $M_{list}$;
    \State Construct state set $S$ and action set $A$;
    \label{code:construct}
    \State Estimate reward $R$ of each $a \in A$;
    \label{code:estimate}
    \For{$i=1$; $i<e_{max}$; $i++$}
      \State Initialize $M_{list} = \varnothing $ and all candidate $M_{list}$ set $M_{all} = \varnothing$;
      \label{code: initialize}
      \State Start state = $s_{0}$, current state $s = s_{0}$;
      \label{code: start state}
      \While{$|M_{list}| < n$}
        \State Initialize the DQN environment using $S$, $A$, and $R$;
        \label{initialize}
        \State Use DQN to find the optimal action $a_{best}$ for $s$;
        \label{use dqn}
        \State $M_{list} \leftarrow M_{list} \cup a_{best}$;
        \label{code: leftarrow}
        \State $s \leftarrow s$ perform $a_{best}$;
        \label{code: perform}
      \EndWhile
      \State $M_{all} \leftarrow M_{all} \cup M_{list}$;
      \label{code; m_all}
    \EndFor \\
    \Return $M_{list} = argmax(M_{all})$;
    \label{code: return}
  \end{algorithmic}
\end{algorithm}

All meta-feature selection steps are summarized in Algorithm \ref{alg:mfselection}. At first, as shown in Line \ref{code:construct}, we construct the state set $S$ and action set $A$, respectively. Then we estimate the reward of each action $R$, which is shown in Line \ref{code:estimate}. The DQN environment is initialized by $S$, $A$, and $R$. For each episode, DQN starts from $S_{0}$, and chooses the maximal reward action in each next step (Line \ref{code: initialize}-\ref{code: perform}). After decoding the termination state, the training results for one episode are obtained. We repeat above steps and eventually obtain the optimal meta-feature list from numerous training results(Line \ref{code: return}).

In the beginning, the lack of experience makes the selection DQN have a deviation from reality.  As the training progresses, DQN will adjust the parameters such as learning rate and discount rate according to the deviation and the selection becomes reasonable. It is just like a human being fixes his action by absorbing the previous experience and the result is getting better.  Eventually, the network parameters become stable and the selected meta-features have the best performance.

With $M_{list}$ and $alg$, all original training datasets can be transformed into a new dataset to train the RF model. Assuming that $|D_{train}| = p$ and $|M_{list}| = y$, we have the new training dataset $D'_{p \times (y + 1)}$, in which the column $D'_{y+1}$ represents the $arg$. After the training phase, our Auto-CASH model works offline. Benefiting from the excellent prediction performance of RF and the high efficiency of GA, the performance of Auto-CASH surpasses early work, which is shown in Section \ref{sec: experimental result}.

\section{Evaluation}\label{sec: eval}

In this section, we evaluate our Auto-CASH approach on the classification CASH problem. Given a dataset, we use Auto-CASH to automatically select an algorithm and search its optimal hyperparameter settings. Then we utilize the new metric criterion in Section~\ref{sec: metric} to examine the performance of results given by Auto-CASH. Eventually, we compare Auto-CASH with classical CASH approach Auto-WEKA and the state-of-the-art CASH approach Auto-Model and discuss the experimental results.

\subsection{Experimental Setup}

For all experiments in this paper, the setup is as follows:

\begin{enumerate}[1)]
\item We implement all experiments in Python 3.7 and run them on a computer with a 2.6 GHz Intel (R) Core (TM) i7-6700HQ CPU and 16 GB RAM.
\item All datasets used are real-world datasets from UCI Machine Learning Repository\footnote{https://archive.ics.uci.edu/ml/index.php} and Kaggle\footnote{https://www.kaggle.com/}. The most significant advantage of using real-world datasets is that it can improve the effect of our model in real life and lay the foundation for future research work. However, for the missing values in the data set, Auto-CASH uses random other values of the same attribute to replace. The implementation of all classification algorithms is from WEKA~\footnote{Source code can refer to https://svn.cms.waikato.ac.nz/svn/weka/branches/stable-3-8/. We wrap the jar package and invoke it using Python.}, which is consistent with Auto-WEKA and Auto-Model.
\item The performance of Auto-WEKA and Auto-Model are both related to the tuning time, so we set the {\itshape timeLimit} parameter to 5 minutes.
\item When calculating the AUC and accuracy value in the metric criterion, we use 80\%  and 20\% of the dataset as the training data and test data, respectively.
\item AUC is the evaluating indicator defined in the binary classification problem. For multiple classification problems, we binarize the output of the classification algorithm using the function in Equation (\ref{equation 4}).

\begin{equation}
\label{equation 4}
f(output)=
\begin{cases}
0& \text{Correct classification}\\
1& \text{Wrong classification}
\end{cases}
\end{equation}

\end{enumerate}

\subsection{Algorithm and Meta-feature Candidates}
\begin{figure}[h]
\centering  
\subfigure[GA tuning curve for K]{
\label{RFpara.sub.1}
\includegraphics[width=0.45\linewidth]{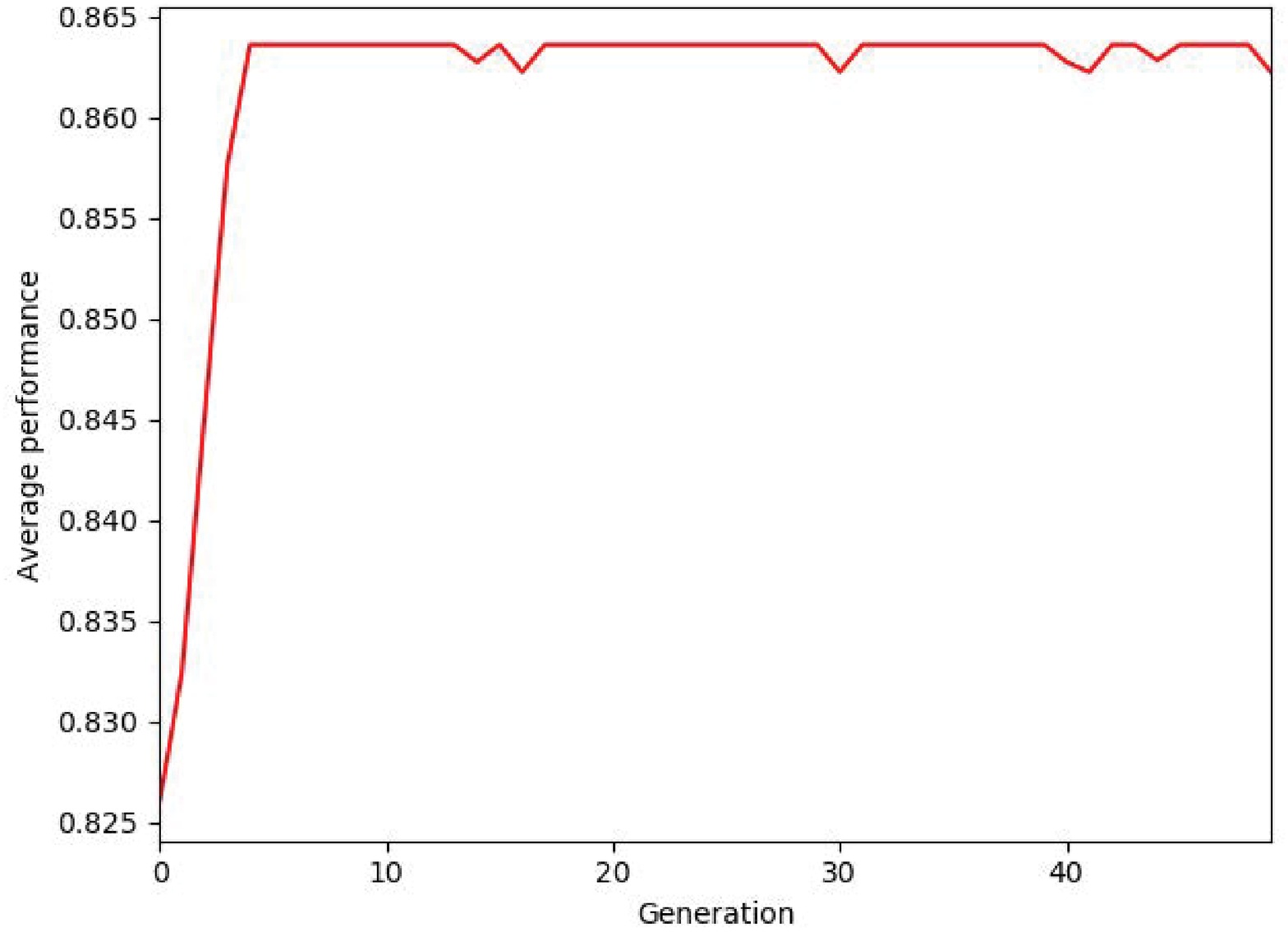}}
\subfigure[GA tuning curve for depth]{
\label{RFpara.sub.2}
\includegraphics[width=0.45\linewidth]{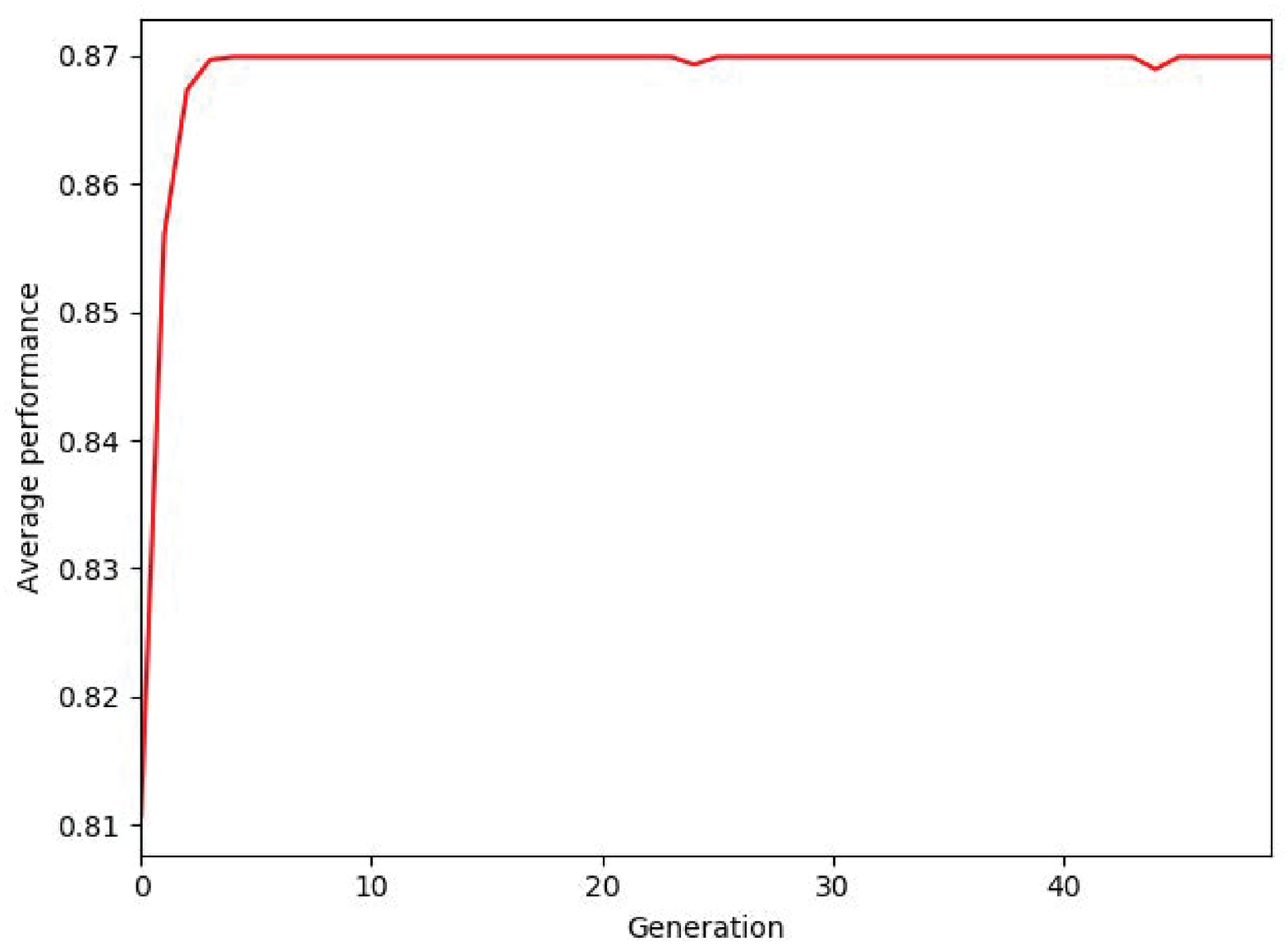}}
\subfigure[GA tuning curve for I]{
\label{RFpara.sub.3}
\includegraphics[width=0.45\linewidth]{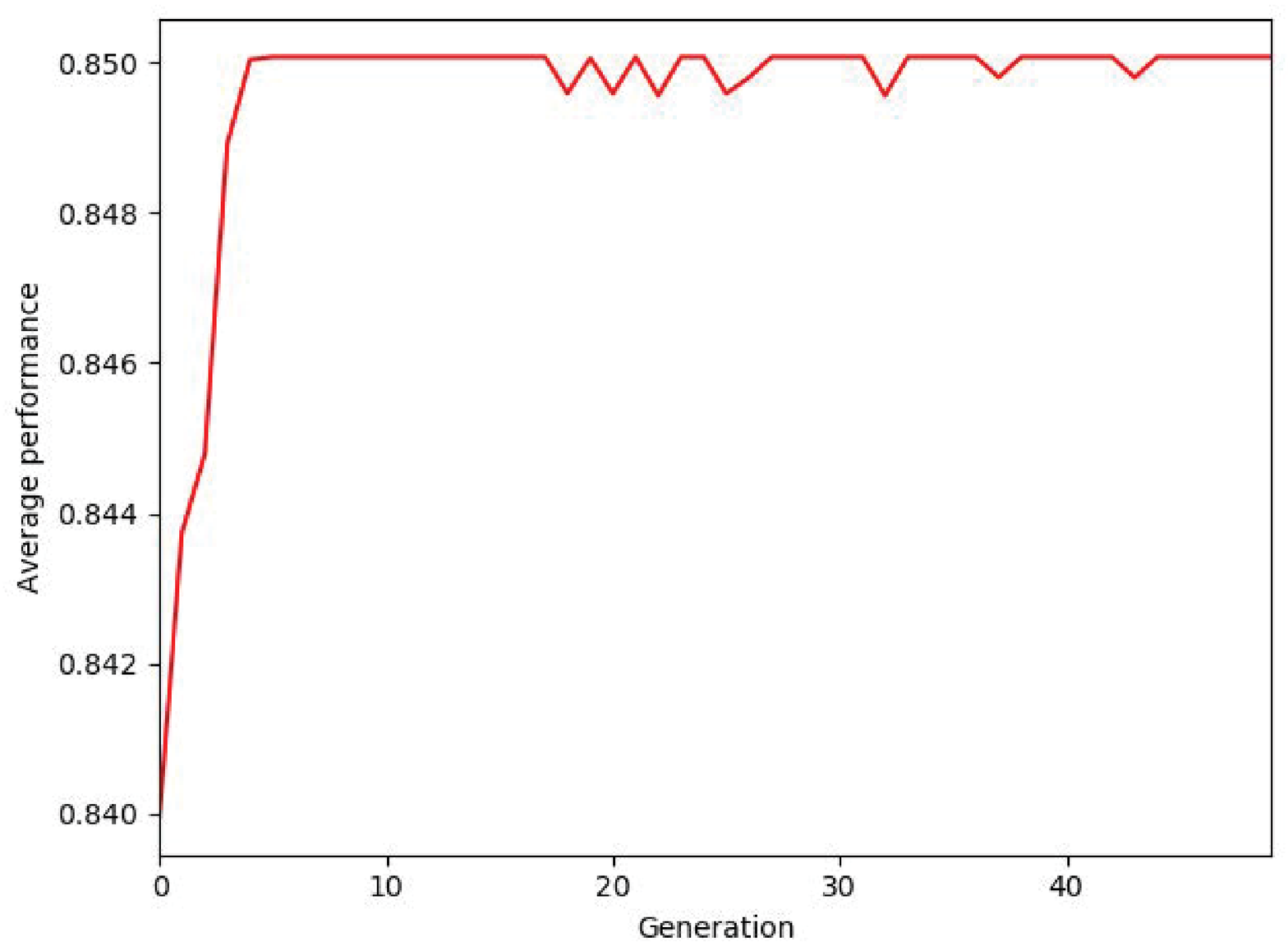}}
\subfigure[Improvement for different hyperparameter]{
\label{RFpara.sub.4}
\includegraphics[width=0.45\linewidth]{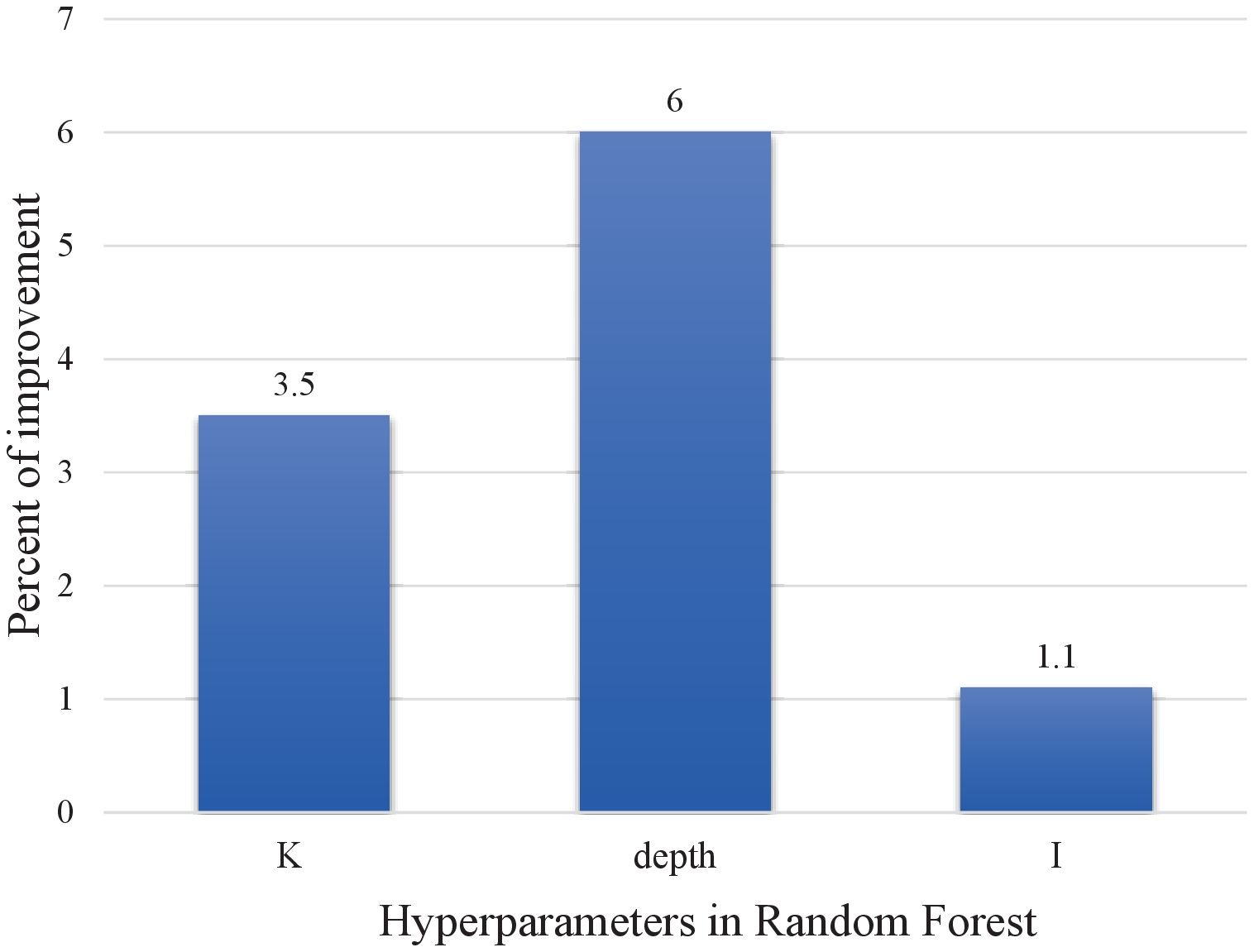}}
\caption{Examples for selecting hyperparameters used in HPO process.}
\label{RFpara.main}
\Description{6666666}
\end{figure}

Referring to the methodology in Sectioni~\ref{sec: hpo}, we first test the performance improvement of hyperparameters for each $alg$. Examples for Random Forest algorithm and ecoli\footnote{https://archive.ics.uci.edu/ml/datasets/Ecoli} dataset is shown in figure \ref{RFpara.main}. Figure \ref{RFpara.sub.1}, \ref{RFpara.sub.2}, and \ref{RFpara.sub.3} represents the GA tuning curve for hyperparameter $-K$, $-depth$, and $-I$, respectively. The x-axis represents each generation in GA, and the y-axis represents the average $F_{score}$ value (performance) of each generation. Although these curves converge in about the fifth generation, the effect of each parameter on the final performance improvement is different, which is shown in Figure~\ref{RFpara.sub.4}. After tuning $-depth$, we can have a $6$ percent improvement, while $-I$ can only improve $1.1$ percent. Thus for $alg$ RF, we decide to tune $-depth$ and $-K$ in HPO process. Table~\ref{tab:hyper-num} shows the number of hyperparameter that needs to be tuned for each algorithm in Auto-CASH.


\begin{table}
  \caption{The number of hyperparameter need to be tuned for each algorithm in Auto-CASH. We totally utilize 23 famous and effective classification algorithms.}
  \label{tab:hyper-num}
  \begin{tabular}{lc|lc}
    \toprule
    Algorithm&Number&Algorithm&Number\\
    \hline
    AdaBoost & 3&Bagging & 3\\
    AttributeSelectedClassifier & 2&BayesNet & 1\\
    ClassificationViaRegression & 2&IBK & 4\\
    DecisionTable & 2&J48 & 8\\
    JRip & 4&KStar & 2\\
    Logistic & 1&LogitBoost & 3\\
    LWL & 3&MultiClass & 3\\
    MultilayerPerceptron & 5&NaiveBayes & 2\\
    RandomCommittee & 2&RandomForest & 2\\
    RandomSubSpace & 3&RandomTree&4\\
    SMO & 6&Vote & 1\\
    LMT & 5& &\\
  \bottomrule
\end{tabular}
\end{table}

\begin{figure}[h]
  \centering
  \includegraphics[width=\linewidth]{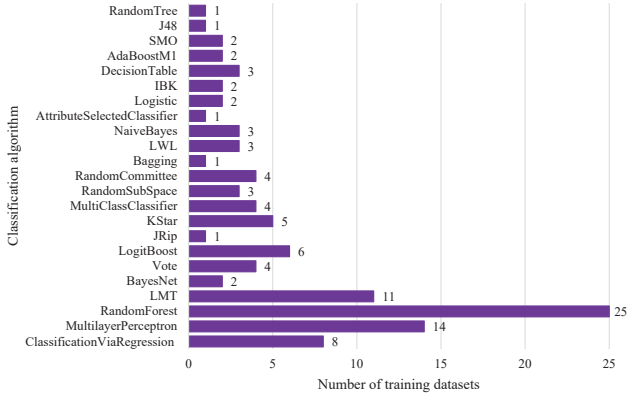}
  \caption{Distribution of the optimal algorithm on 104 training datasets. For each algorithm, we list the number of datasets using it as its optimal algorithm.}
  \Description{77777777}
  \label{algodistribute pic}
\end{figure}

After selecting the hyperparameters to be tuned in HPO, we test the optimal algorithm for each dataset in $D_{train}$. Then we compare the performance of all algorithm candidates on training datasets and list their optimal algorithm in Figure~\ref{algodistribute pic}.

Meta-features used for representing a dataset in our experiments are summerized as follows:
\begin{itemize}
  \item{$mf_{0}$}: Class number in target attribute.
  \item{$mf_{1}$}: Class information entropy of target attribute.
  \item{$mf_{2}$}: Maximum proportion of single class in target attribute.
  \item{$mf_{3}$}: Minimum proportion of single class in target attribute.
  \item{$mf_{4}$}: Number of numeric attributes.
  \item{$mf_{5}$}: Number of category attributes.
  \item{$mf_{6}$}: Proportion of numeric attributes.
  \item{$mf_{7}$}: Total number of attributes.
  \item{$mf_{8}$}: Records number in the dataset.
  \item{$mf_{9}$}: Class number in category attribute with the least classes.
  \item{$mf_{10}$}: Class information entropy in category attribute with the least classes.
  \item{$mf_{11}$}: Maximum proportion of single class in category attribute with the least classes.
  \item{$mf_{12}$}: Minimum proportion of single class in category attribute with the least classes.
  \item{$mf_{13}$}: Class number in category attribute with the most classes.
  \item{$mf_{14}$}: Class information entropy in category attribute with the most classes.
  \item{$mf_{15}$}: Maximum proportion of single class in category attribute with the most classes.
  \item{$mf_{16}$}: Minimum proportion of single class in category attribute with the most classes.
  \item{$mf_{17}$}: Minimum average value in numeric attributes.
  \item{$mf_{18}$}: Maximum average value in numeric attributes.
  \item{$mf_{19}$}: Minimum variance in numeric attributes.
  \item{$mf_{20}$}: Maximum variance in numeric attributes.
  \item{$mf_{21}$}: Variance of average value in numeric attributes.
  \item{$mf_{22}$}: Variance of variance in numeric attributes.
\end{itemize}
The type mentioned in Section~\ref{sec: methodology} of each $mf$ is shown in Table~\ref{tab:type of each mf}. These meta-features are easy to calculate, which will reduce calculation cost in the algorithm selection.

\begin{table}
  \caption{Type of each meta-feature.}
  \label{tab:type of each mf}
  \begin{tabular}{cc}
    \toprule
    Type&$mf$ index\\
    \midrule
    Type 1& 1, 10, 14\\
    Type 2& 2, 3, 6, 11, 12, 15, 16\\
    Type 3& 17, 18\\
    Type 4& 19, 20, 21, 22\\
    Type 5& 0, 4, 5, 7, 8, 9, 13\\
  \bottomrule
\end{tabular}
\end{table}

\subsection{Experimental results}
\label{sec: experimental result}
\begin{figure}[h]
  \centering
  \includegraphics[width=\linewidth]{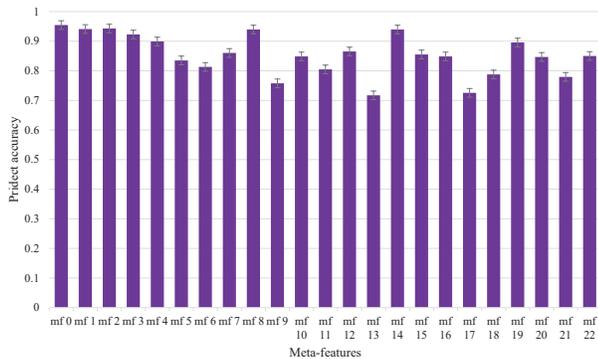}
  \caption{Performance for each meta-feature.}
  \Description{8888888}
  \label{mfperform pic}
\end{figure}

After determining $Alg$ and $MF$, we utilize DQN to obtain $M_{list}$. Too many meta-features will not bring enough information gain while increasing the computational complexity. Therefore, we set the upper limit of $|M_{list}|$ to 8 and evaluate each $mf$ with different batch sizes range from 2 to 8.  The evaluation results are shown in Figure~\ref{mfperform pic}, which represents the estimated reward of $mf$. From the results, we can see that the influence of each $mf$ has a large range. According to the methodology in Section~\ref{sec: methodology}, there is totally 23 actions and $2^{23}$ states. The experience memory size of DQN is set to 200, which will be randomly updated after an action decision. Then we get the $M_{list} = \{mf_{0}, mf_{2}, mf_{4}, mf_{6}, mf_{7}, mf_{9}, mf_{13}\}$ among the outputs of DQN. We utilize these selected meta-features and each dataset's optimal algorithm to train the RF. The trained RF will predict the optimal algorithm for test datasets. Eventually in HPO, we use GA and set the maximum generations to 50.

We evaluate the $F_{score}$ performance of Auto-CASH on 20 classification datasets in Table~\ref{tab:data information}. The average time cost in each phase is shown in table \ref{tab:time cost}. From the table, we can see that Auto-CASH costs few time on autonomous algorithm selection. After tuning for hyperparameter, the HPO time is greatly reduced, which guarantees the efficiency of Auto-CASH. We also evaluate the $F_{score}$performance of Auto-WEKA and Auto-Model on the same datasets, and the detailed expermental results are shown in Table~\ref{tab:fscore}.

\begin{table}
  \caption{Datasets}
  \label{tab:data information}
  \begin{tabular}{ccccc}
    \toprule
    Dataset&Records&Attributes&Classes&Symbol\\
    \midrule
    Avila& 20867&10&12&$D_{1}$\\
    Nursery&12960&8&3&$D_{2}$\\
    Absenteeism&740&21&36&$D_{3}$\\
    Climate&540&19&2&$D_{4}$\\
    Australian&690&14&2&$D_{5}$\\
    Iris.2D&150&2&3&$D_{6}$\\
    Heart-c&303&14&5&$D_{7}$\\
    Sick&3772&30&2&$D_{8}$\\
    Anneal&798&38&6&$D_{9}$\\
    Hypothyroid&3772&27&2&$D_{10}$\\
    Squash&52&24&3&$D_{11}$\\
    Vowel&990&14&11&$D_{12}$\\
    Zoo&101&18&7&$D_{13}$\\
    Breast-W&699&9&2&$D_{14}$\\
    Iris&150&4&3&$D_{15}$\\
    Diabetes&768&9&2&$D_{16}$\\
    Dermatology&336&34&6&$D_{17}$\\
    Musk&476&166&2&$D_{18}$\\
    Promoter&106&57&2&$D_{19}$\\
    Blood&748&5&2&$D_{20}$\\
  \bottomrule
\end{tabular}
\end{table}

\begin{table}
  \caption{The average time of each phase in Auto-CASH.}
  \label{tab:time cost}
  \begin{tabular}{cc}
    \toprule
    Phase&Time\\
    \midrule
    DQN training&10 CPU hour\\
    Calculate meta-feature value & 0.96 second\\
    Algorithm selection&  0.5 second\\
    HPO& 229.3 seconds\\
    Total CASH& 230.76 seconds\\
  \bottomrule
\end{tabular}
\end{table}

\begin{table}
  \caption{$F_{score}$ of Auto-CASH, Auto-Model, and Auto-WEKA on test datasets. Bold font denotes the best result. Auto-Model cannot give a result for some cases, so we use -1 here.}
  \label{tab:fscore}
  \begin{tabular}{cccc}
    \toprule
    Dataset&Auto-CASH&Auto-Model&Auto-WEKA\\
    \midrule
    $D_{1}$& \textbf{0.998}&0.987&0.996\\
    $D_{2}$& \textbf{0.996}&0.942&0.947\\
    $D_{3}$& \textbf{0.408}&0.363&0.36\\
    $D_{4}$& 0.925&\textbf{0.948}&0.711\\
    $D_{5}$& \textbf{0.845}&0.806&0.790\\
    $D_{6}$& 0.967&0.965&\textbf{1.0}\\
    $D_{7}$& \textbf{0.882}&-1&0.58\\
    $D_{8}$& \textbf{0.978}&-1&0.886\\
    $D_{9}$& 0.969&0.38&\textbf{0.974}\\
    $D_{10}$& \textbf{0.988}&-1&0.976\\
    $D_{11}$& \textbf{0.563}&0.409&0.509\\
    $D_{12}$& \textbf{0.963}&0.591&0.11\\
    $D_{13}$& \textbf{1.0}&0.9&0.569\\
    $D_{14}$& \textbf{0.961}&0.957&0.952\\
    $D_{15}$& \textbf{0.979}&0.964&0.966\\
    $D_{16}$& 0.677&\textbf{0.686}&0.633\\
    $D_{17}$& \textbf{0.986} &-1&0.942\\
    $D_{18}$& \textbf{0.951}&\textbf{0.951}&\textbf{0.951}\\
    $D_{19}$& \textbf{0.952}&0.697&0.935\\
    $D_{20}$&\textbf{0.611}&0.569&0.478\\
  \bottomrule
\end{tabular}
\end{table}

\subsection{Discussion}
The meta-feature selected by DQN can comprehensively represent the datasets. Compared with Auto-Model, we use fewer meta-features while Auto-CASH achieves a better performance in most cases as shown in Table~\ref{tab:fscore}. It proves that DQN is more effective. Our approach significantly reduces human labor in the training phase, which makes it a fully-automated model. Auto-CASH can handle data missing anomalies, which makes it more robust for various dataset than Auto-Model.

Auto-CASH achieves better performance in shorter time. We first evaluate the hyperparameters for each algorithm and finally select some of them to tune in HPO process. The results in Table~\ref{tab:time cost} demonstrates that it is meaningful and efficient. Reducing the complexity of the hyperparameter space means that the optimal result can be found in shorter time. RF also made crucial contributions in reducing time, which is the advantage of the pre-trained model. Compared to Auto-WEKA and Auto-Model, we save about a quarter of time cost while obtaining the same or better results(5 minutes for Auto-WEKA and Auto-Model). It has a significantly meaning in such era of explosive data growth.

Overall, the design of Auto-CASH is reasonable and meaningful. Auto-CASH can utilize the experience learned before to give better results for new tasks within a shorter time. It outperforms the state-of-the-art Auto-Model and classical Auto-WEKA.

\section{Conclusion and future work}
\label{sec: conclusion}
In this paper, we present Auto-CASH, a pre-trained model based on meta-learning for the CASH problem. By transforming the selection of meta-feature into a continuous action decision problem, we are able to automatically solve it utilizing Deep Q-Network. Thus it significantly reduces human labor in the training process. For a particular task, Auto-CASH enhances the performance of the recommended algorithm within an acceptable time by means of Random Forest and Genetic Algorithm. Experimental results demonstrate that Auto-CASH outperforms classical and the state-of-the-art CASH approach on efficiency and effectiveness. In future work, we plan to extend Auto-CASH to deal with more problems, e.g., regression, image processing. Besides, we intend to develop an approach to automatically extract the meta-feature candidates according to the task and its datasets.


\begin{thebibliography}{27}


\ifx \showCODEN    \undefined \def \showCODEN     #1{\unskip}     \fi
\ifx \showDOI      \undefined \def \showDOI       #1{#1}\fi
\ifx \showISBNx    \undefined \def \showISBNx     #1{\unskip}     \fi
\ifx \showISBNxiii \undefined \def \showISBNxiii  #1{\unskip}     \fi
\ifx \showISSN     \undefined \def \showISSN      #1{\unskip}     \fi
\ifx \showLCCN     \undefined \def \showLCCN      #1{\unskip}     \fi
\ifx \shownote     \undefined \def \shownote      #1{#1}          \fi
\ifx \showarticletitle \undefined \def \showarticletitle #1{#1}   \fi
\ifx \showURL      \undefined \def \showURL       {\relax}        \fi
\providecommand\bibfield[2]{#2}
\providecommand\bibinfo[2]{#2}
\providecommand\natexlab[1]{#1}
\providecommand\showeprint[2][]{arXiv:#2}

\bibitem[\protect\citeauthoryear{Bergstra and Bengio}{Bergstra and
  Bengio}{2012}]%
        {bergstra2012random}
\bibfield{author}{\bibinfo{person}{James Bergstra} {and}
  \bibinfo{person}{Yoshua Bengio}.} \bibinfo{year}{2012}\natexlab{}.
\newblock \showarticletitle{Random search for hyper-parameter optimization}.
\newblock \bibinfo{journal}{\emph{Journal of machine learning research}}
  \bibinfo{volume}{13}, \bibinfo{number}{Feb} (\bibinfo{year}{2012}),
  \bibinfo{pages}{281--305}.
\newblock


\bibitem[\protect\citeauthoryear{Bilalli, Abell{\'o}, and Aluja-Banet}{Bilalli
  et~al\mbox{.}}{2017}]%
        {bilalli2017predictive}
\bibfield{author}{\bibinfo{person}{Besim Bilalli}, \bibinfo{person}{Alberto
  Abell{\'o}}, {and} \bibinfo{person}{Tomas Aluja-Banet}.}
  \bibinfo{year}{2017}\natexlab{}.
\newblock \showarticletitle{On the predictive power of meta-features in
  OpenML}.
\newblock \bibinfo{journal}{\emph{International Journal of Applied Mathematics
  and Computer Science}} \bibinfo{volume}{27}, \bibinfo{number}{4}
  (\bibinfo{year}{2017}), \bibinfo{pages}{697--712}.
\newblock


\bibitem[\protect\citeauthoryear{Brochu, Cora, and De~Freitas}{Brochu
  et~al\mbox{.}}{2010}]%
        {brochu2010tutorial}
\bibfield{author}{\bibinfo{person}{Eric Brochu}, \bibinfo{person}{Vlad~M Cora},
  {and} \bibinfo{person}{Nando De~Freitas}.} \bibinfo{year}{2010}\natexlab{}.
\newblock \showarticletitle{A tutorial on Bayesian optimization of expensive
  cost functions, with application to active user modeling and hierarchical
  reinforcement learning}.
\newblock \bibinfo{journal}{\emph{arXiv preprint arXiv:1012.2599}}
  (\bibinfo{year}{2010}).
\newblock


\bibitem[\protect\citeauthoryear{Dahl, Sainath, and Hinton}{Dahl
  et~al\mbox{.}}{2013}]%
        {dahl2013improving}
\bibfield{author}{\bibinfo{person}{George~E Dahl}, \bibinfo{person}{Tara~N
  Sainath}, {and} \bibinfo{person}{Geoffrey~E Hinton}.}
  \bibinfo{year}{2013}\natexlab{}.
\newblock \showarticletitle{Improving deep neural networks for LVCSR using
  rectified linear units and dropout}. In \bibinfo{booktitle}{\emph{2013 IEEE
  international conference on acoustics, speech and signal processing}}. IEEE,
  \bibinfo{pages}{8609--8613}.
\newblock


\bibitem[\protect\citeauthoryear{Fawcett}{Fawcett}{2006}]%
        {fawcett2006introduction}
\bibfield{author}{\bibinfo{person}{Tom Fawcett}.}
  \bibinfo{year}{2006}\natexlab{}.
\newblock \showarticletitle{An introduction to ROC analysis}.
\newblock \bibinfo{journal}{\emph{Pattern recognition letters}}
  \bibinfo{volume}{27}, \bibinfo{number}{8} (\bibinfo{year}{2006}),
  \bibinfo{pages}{861--874}.
\newblock


\bibitem[\protect\citeauthoryear{Filchenkov and Pendryak}{Filchenkov and
  Pendryak}{2015}]%
        {filchenkov2015datasets}
\bibfield{author}{\bibinfo{person}{Andrey Filchenkov} {and}
  \bibinfo{person}{Arseniy Pendryak}.} \bibinfo{year}{2015}\natexlab{}.
\newblock \showarticletitle{Datasets meta-feature description for recommending
  feature selection algorithm}. In \bibinfo{booktitle}{\emph{2015 Artificial
  Intelligence and Natural Language and Information Extraction, Social Media
  and Web Search FRUCT Conference (AINL-ISMW FRUCT)}}. IEEE,
  \bibinfo{pages}{11--18}.
\newblock


\bibitem[\protect\citeauthoryear{Hutter, Kotthoff, and Vanschoren}{Hutter
  et~al\mbox{.}}{2019}]%
        {hutter2019automated}
\bibfield{author}{\bibinfo{person}{Frank Hutter}, \bibinfo{person}{Lars
  Kotthoff}, {and} \bibinfo{person}{Joaquin Vanschoren}.}
  \bibinfo{year}{2019}\natexlab{}.
\newblock \bibinfo{booktitle}{\emph{Automated Machine Learning}}.
\newblock \bibinfo{publisher}{Springer}.
\newblock


\bibitem[\protect\citeauthoryear{Lake, Ullman, Tenenbaum, and Gershman}{Lake
  et~al\mbox{.}}{2017}]%
        {lake2017building}
\bibfield{author}{\bibinfo{person}{Brenden~M Lake}, \bibinfo{person}{Tomer~D
  Ullman}, \bibinfo{person}{Joshua~B Tenenbaum}, {and}
  \bibinfo{person}{Samuel~J Gershman}.} \bibinfo{year}{2017}\natexlab{}.
\newblock \showarticletitle{Building machines that learn and think like
  people}.
\newblock \bibinfo{journal}{\emph{Behavioral and brain sciences}}
  \bibinfo{volume}{40} (\bibinfo{year}{2017}).
\newblock


\bibitem[\protect\citeauthoryear{Li, Jamieson, DeSalvo, Rostamizadeh, and
  Talwalkar}{Li et~al\mbox{.}}{2017}]%
        {li2017hyperband}
\bibfield{author}{\bibinfo{person}{Lisha Li}, \bibinfo{person}{Kevin Jamieson},
  \bibinfo{person}{Giulia DeSalvo}, \bibinfo{person}{Afshin Rostamizadeh},
  {and} \bibinfo{person}{Ameet Talwalkar}.} \bibinfo{year}{2017}\natexlab{}.
\newblock \showarticletitle{Hyperband: A novel bandit-based approach to
  hyperparameter optimization}.
\newblock \bibinfo{journal}{\emph{The Journal of Machine Learning Research}}
  \bibinfo{volume}{18}, \bibinfo{number}{1} (\bibinfo{year}{2017}),
  \bibinfo{pages}{6765--6816}.
\newblock


\bibitem[\protect\citeauthoryear{Lindauer, van Rijn, and Kotthoff}{Lindauer
  et~al\mbox{.}}{2019}]%
        {lindauer2019algorithm}
\bibfield{author}{\bibinfo{person}{Marius Lindauer}, \bibinfo{person}{Jan~N van
  Rijn}, {and} \bibinfo{person}{Lars Kotthoff}.}
  \bibinfo{year}{2019}\natexlab{}.
\newblock \showarticletitle{The algorithm selection competitions 2015 and
  2017}.
\newblock \bibinfo{journal}{\emph{Artificial Intelligence}}
  \bibinfo{volume}{272} (\bibinfo{year}{2019}), \bibinfo{pages}{86--100}.
\newblock


\bibitem[\protect\citeauthoryear{Melo}{Melo}{2001}]%
        {melo2001convergence}
\bibfield{author}{\bibinfo{person}{Francisco~S Melo}.}
  \bibinfo{year}{2001}\natexlab{}.
\newblock \showarticletitle{Convergence of Q-learning: A simple proof}.
\newblock \bibinfo{journal}{\emph{Institute Of Systems and Robotics, Tech.
  Rep}} (\bibinfo{year}{2001}), \bibinfo{pages}{1--4}.
\newblock


\bibitem[\protect\citeauthoryear{Mnih, Kavukcuoglu, Silver, Graves, Antonoglou,
  Wierstra, and Riedmiller}{Mnih et~al\mbox{.}}{2013}]%
        {mnih2013playing}
\bibfield{author}{\bibinfo{person}{Volodymyr Mnih}, \bibinfo{person}{Koray
  Kavukcuoglu}, \bibinfo{person}{David Silver}, \bibinfo{person}{Alex Graves},
  \bibinfo{person}{Ioannis Antonoglou}, \bibinfo{person}{Daan Wierstra}, {and}
  \bibinfo{person}{Martin Riedmiller}.} \bibinfo{year}{2013}\natexlab{}.
\newblock \showarticletitle{Playing atari with deep reinforcement learning}.
\newblock \bibinfo{journal}{\emph{arXiv preprint arXiv:1312.5602}}
  (\bibinfo{year}{2013}).
\newblock


\bibitem[\protect\citeauthoryear{Montgomery}{Montgomery}{2017}]%
        {montgomery2017design}
\bibfield{author}{\bibinfo{person}{Douglas~C Montgomery}.}
  \bibinfo{year}{2017}\natexlab{}.
\newblock \bibinfo{booktitle}{\emph{Design and analysis of experiments}}.
\newblock \bibinfo{publisher}{John wiley \& sons}.
\newblock


\bibitem[\protect\citeauthoryear{Olson and Moore}{Olson and Moore}{2019}]%
        {olson2019tpot}
\bibfield{author}{\bibinfo{person}{Randal~S Olson} {and}
  \bibinfo{person}{Jason~H Moore}.} \bibinfo{year}{2019}\natexlab{}.
\newblock \showarticletitle{TPOT: A tree-based pipeline optimization tool for
  automating machine learning}.
\newblock In \bibinfo{booktitle}{\emph{Automated Machine Learning}}.
  \bibinfo{publisher}{Springer}, \bibinfo{pages}{151--160}.
\newblock


\bibitem[\protect\citeauthoryear{Pelikan, Goldberg, Cant{\'u}-Paz,
  et~al\mbox{.}}{Pelikan et~al\mbox{.}}{1999}]%
        {pelikan1999boa}
\bibfield{author}{\bibinfo{person}{Martin Pelikan}, \bibinfo{person}{David~E
  Goldberg}, \bibinfo{person}{Erick Cant{\'u}-Paz}, {et~al\mbox{.}}}
  \bibinfo{year}{1999}\natexlab{}.
\newblock \showarticletitle{BOA: The Bayesian optimization algorithm}. In
  \bibinfo{booktitle}{\emph{Proceedings of the genetic and evolutionary
  computation conference GECCO-99}}, Vol.~\bibinfo{volume}{1}.
  \bibinfo{pages}{525--532}.
\newblock


\bibitem[\protect\citeauthoryear{Powers}{Powers}{2011}]%
        {powers2011evaluation}
\bibfield{author}{\bibinfo{person}{David~Martin Powers}.}
  \bibinfo{year}{2011}\natexlab{}.
\newblock \showarticletitle{Evaluation: from precision, recall and F-measure to
  ROC, informedness, markedness and correlation}.
\newblock  (\bibinfo{year}{2011}).
\newblock


\bibitem[\protect\citeauthoryear{Schaffer}{Schaffer}{1994}]%
        {schaffer1994cross}
\bibfield{author}{\bibinfo{person}{Cullen Schaffer}.}
  \bibinfo{year}{1994}\natexlab{}.
\newblock \showarticletitle{Cross-validation, stacking and bi-level stacking:
  Meta-methods for classification learning}.
\newblock In \bibinfo{booktitle}{\emph{Selecting Models from Data}}.
  \bibinfo{publisher}{Springer}, \bibinfo{pages}{51--59}.
\newblock


\bibitem[\protect\citeauthoryear{Snoek, Larochelle, and Adams}{Snoek
  et~al\mbox{.}}{2012}]%
        {snoek2012practical}
\bibfield{author}{\bibinfo{person}{Jasper Snoek}, \bibinfo{person}{Hugo
  Larochelle}, {and} \bibinfo{person}{Ryan~P Adams}.}
  \bibinfo{year}{2012}\natexlab{}.
\newblock \showarticletitle{Practical bayesian optimization of machine learning
  algorithms}. In \bibinfo{booktitle}{\emph{Advances in neural information
  processing systems}}. \bibinfo{pages}{2951--2959}.
\newblock


\bibitem[\protect\citeauthoryear{Snoek, Rippel, Swersky, Kiros, Satish,
  Sundaram, Patwary, Prabhat, and Adams}{Snoek et~al\mbox{.}}{2015}]%
        {snoek2015scalable}
\bibfield{author}{\bibinfo{person}{Jasper Snoek}, \bibinfo{person}{Oren
  Rippel}, \bibinfo{person}{Kevin Swersky}, \bibinfo{person}{Ryan Kiros},
  \bibinfo{person}{Nadathur Satish}, \bibinfo{person}{Narayanan Sundaram},
  \bibinfo{person}{Mostofa Patwary}, \bibinfo{person}{Mr Prabhat}, {and}
  \bibinfo{person}{Ryan Adams}.} \bibinfo{year}{2015}\natexlab{}.
\newblock \showarticletitle{Scalable bayesian optimization using deep neural
  networks}. In \bibinfo{booktitle}{\emph{International conference on machine
  learning}}. \bibinfo{pages}{2171--2180}.
\newblock


\bibitem[\protect\citeauthoryear{Taylor, Marco, Wolff, Elkhatib, and
  Wang}{Taylor et~al\mbox{.}}{2018}]%
        {taylor2018adaptive}
\bibfield{author}{\bibinfo{person}{Ben Taylor}, \bibinfo{person}{Vicent~Sanz
  Marco}, \bibinfo{person}{Willy Wolff}, \bibinfo{person}{Yehia Elkhatib},
  {and} \bibinfo{person}{Zheng Wang}.} \bibinfo{year}{2018}\natexlab{}.
\newblock \showarticletitle{Adaptive deep learning model selection on embedded
  systems}.
\newblock \bibinfo{journal}{\emph{ACM SIGPLAN Notices}} \bibinfo{volume}{53},
  \bibinfo{number}{6} (\bibinfo{year}{2018}), \bibinfo{pages}{31--43}.
\newblock


\bibitem[\protect\citeauthoryear{Thornton, Hutter, Hoos, and
  Leyton-Brown}{Thornton et~al\mbox{.}}{2013}]%
        {thornton2013auto}
\bibfield{author}{\bibinfo{person}{Chris Thornton}, \bibinfo{person}{Frank
  Hutter}, \bibinfo{person}{Holger~H Hoos}, {and} \bibinfo{person}{Kevin
  Leyton-Brown}.} \bibinfo{year}{2013}\natexlab{}.
\newblock \showarticletitle{Auto-WEKA: Combined selection and hyperparameter
  optimization of classification algorithms}. In
  \bibinfo{booktitle}{\emph{Proceedings of the 19th ACM SIGKDD international
  conference on Knowledge discovery and data mining}}.
  \bibinfo{pages}{847--855}.
\newblock


\bibitem[\protect\citeauthoryear{Wang, Wang, Mu, Li, and Gao}{Wang
  et~al\mbox{.}}{2019}]%
        {wang2019auto}
\bibfield{author}{\bibinfo{person}{Chunnan Wang}, \bibinfo{person}{Hongzhi
  Wang}, \bibinfo{person}{Tianyu Mu}, \bibinfo{person}{Jianzhong Li}, {and}
  \bibinfo{person}{Hong Gao}.} \bibinfo{year}{2019}\natexlab{}.
\newblock \showarticletitle{Auto-Model: Utilizing Research Papers and HPO
  Techniques to Deal with the CASH problem}.
\newblock \bibinfo{journal}{\emph{arXiv preprint arXiv:1910.10902}}
  (\bibinfo{year}{2019}).
\newblock


\bibitem[\protect\citeauthoryear{Watkins and Dayan}{Watkins and Dayan}{1992}]%
        {watkins1992q}
\bibfield{author}{\bibinfo{person}{Christopher~JCH Watkins} {and}
  \bibinfo{person}{Peter Dayan}.} \bibinfo{year}{1992}\natexlab{}.
\newblock \showarticletitle{Q-learning}.
\newblock \bibinfo{journal}{\emph{Machine learning}} \bibinfo{volume}{8},
  \bibinfo{number}{3-4} (\bibinfo{year}{1992}), \bibinfo{pages}{279--292}.
\newblock


\bibitem[\protect\citeauthoryear{Whitley}{Whitley}{1994}]%
        {whitley1994genetic}
\bibfield{author}{\bibinfo{person}{Darrell Whitley}.}
  \bibinfo{year}{1994}\natexlab{}.
\newblock \showarticletitle{A genetic algorithm tutorial}.
\newblock \bibinfo{journal}{\emph{Statistics and computing}}
  \bibinfo{volume}{4}, \bibinfo{number}{2} (\bibinfo{year}{1994}),
  \bibinfo{pages}{65--85}.
\newblock


\bibitem[\protect\citeauthoryear{Xiao, Rasul, and Vollgraf}{Xiao
  et~al\mbox{.}}{2017}]%
        {xiao2017fashion}
\bibfield{author}{\bibinfo{person}{Han Xiao}, \bibinfo{person}{Kashif Rasul},
  {and} \bibinfo{person}{Roland Vollgraf}.} \bibinfo{year}{2017}\natexlab{}.
\newblock \showarticletitle{Fashion-mnist: a novel image dataset for
  benchmarking machine learning algorithms}.
\newblock \bibinfo{journal}{\emph{arXiv preprint arXiv:1708.07747}}
  (\bibinfo{year}{2017}).
\newblock


\bibitem[\protect\citeauthoryear{Xiujuan and Zhongke}{Xiujuan and
  Zhongke}{2004}]%
        {xiujuan2004overview}
\bibfield{author}{\bibinfo{person}{Lei Xiujuan} {and} \bibinfo{person}{Shi
  Zhongke}.} \bibinfo{year}{2004}\natexlab{}.
\newblock \showarticletitle{Overview of multi-objective optimization methods}.
\newblock \bibinfo{journal}{\emph{Journal of Systems Engineering and
  Electronics}} \bibinfo{volume}{15}, \bibinfo{number}{2}
  (\bibinfo{year}{2004}), \bibinfo{pages}{142--146}.
\newblock


\bibitem[\protect\citeauthoryear{Zames, Ajlouni, Ajlouni, Ajlouni, Holland,
  Hills, and Goldberg}{Zames et~al\mbox{.}}{1981}]%
        {zames1981genetic}
\bibfield{author}{\bibinfo{person}{G Zames}, \bibinfo{person}{NM Ajlouni},
  \bibinfo{person}{NM Ajlouni}, \bibinfo{person}{NM Ajlouni},
  \bibinfo{person}{JH Holland}, \bibinfo{person}{WD Hills}, {and}
  \bibinfo{person}{DE Goldberg}.} \bibinfo{year}{1981}\natexlab{}.
\newblock \showarticletitle{Genetic algorithms in search, optimization and
  machine learning.}
\newblock \bibinfo{journal}{\emph{Information Technology Journal}}
  \bibinfo{volume}{3}, \bibinfo{number}{1} (\bibinfo{year}{1981}),
  \bibinfo{pages}{301--302}.
\newblock


\end{thebibliography}
\end{document}